\documentclass[11pt]{article}
\usepackage[margin=1in]{geometry}
\usepackage{amsmath, amssymb}
\usepackage{graphicx}
\usepackage{booktabs}
\usepackage{hyperref}
\usepackage{microtype}
\title{Time Without Timesteps: Simulating Coupled Dynamical Systems via Self-Consistency}
\author{Liyu Zerihun \and Mark Shinyoung Lee}
\date{}
\begin{document}
\maketitle

\begin{abstract}
Numerical simulation of dynamical systems is usually organized as a causal march through time: each state is computed from the previous state, and the full trajectory is obtained by repeated local updates. This paper explores a different computational formulation for coupled dynamical systems. We train neural surrogates that map full input trajectories and initial conditions directly to full output trajectories for individual subsystem types. Building on classical waveform relaxation, we replace the numerical integration performed within each subsystem update with these learned trajectory operators. Coupled systems are then assembled by enforcing self-consistency between the trajectories predicted by each subsystem and those supplied as inputs to its neighbors. Simulation therefore becomes a fixed-point problem over complete trajectories rather than an explicit timestep-by-timestep rollout.

We demonstrate the approach on van der Pol oscillators and Hodgkin--Huxley neuron networks. The sequential depth of the computation becomes the number of solver iterations rather than the number of timesteps: we recover coupled trajectories in $4$--$10$ Newton iterations where the reference integrator takes $1500$ steps. The gradient of the simulation likewise loses its time recursion: it is the solution of a linear system, not a backward pass through the integrator, and we solve it by GMRES at memory independent of solver depth. We show that a single scalar measured from the learned operator---the spectral radius of its Jacobian---predicts in advance where the coupled solve will converge, and that this prediction holds across both coupling strength and stiffness. We also identify a regime in which the implicit gradient is the only one available: past the contraction boundary, unrolled backpropagation diverges and a Neumann adjoint fails, while the implicit gradient remains correct to $0.04\%$ against finite differences.

The method is not presented as a replacement for classical solvers. It is an early exploration of a trajectory-level simulation paradigm in which learned module operators are composed through self-consistency. We report regimes in which the approach succeeds and regimes in which surrogate error causes degradation, and we identify surrogate accuracy, fixed-point conditioning, and training-distribution coverage as the main practical bottlenecks.
\end{abstract}

\section{Introduction}

To simulate a differential equation, we almost always march: given the current state, a numerical integrator estimates the next state, and repeating this local update produces a trajectory.

That procedure is powerful, mature, and often the right tool. But it is not the only way to view the computation. A differential equation, together with its initial or boundary conditions, defines a set of admissible trajectories, and the physical solution is the one that satisfies the dynamics and all coupling constraints simultaneously. In this sense the trajectory is a global object selected by consistency; the order in which a solver happens to construct it is incidental.

This distinction matters most in coupled systems. When subsystems interact, each component's trajectory depends on signals produced by its neighbors. If those driving signals were known, each subsystem could be solved independently. The difficulty is circular: the driving signals are themselves functions of the unknown trajectories. Standard solvers resolve this circularity locally in time by advancing all coupled states together. We instead ask whether the circularity can be resolved globally, at the level of whole trajectories.

That question has an established answer in numerical analysis. Waveform relaxation, introduced for circuit simulation \cite{lelarasmee1982waveform}, solves exactly this fixed point: each subsystem is integrated over the whole time window under an assumed input waveform, and the process is repeated until the waveforms agree. The method never became a general-purpose tool, and one reason is cost. Every relaxation sweep requires a full numerical integration of every subsystem, so each iteration is as expensive as a complete simulation, and the derivative of a sweep is expensive for the same reason.

\emph{Time Without Timesteps} (TWT) asks what changes when that inner integration is replaced by a learned trajectory operator. For each subsystem type, we train a neural surrogate that maps a full driving input trajectory and initial condition to a full output trajectory. A coupled system is then assembled by requiring mutual agreement: each subsystem's output must equal the trajectory predicted by its surrogate under the inputs induced by neighboring subsystems.

What motivates the construction is the structure of the computation, not raw speed.

One consequence is sequential depth. A conventional solve is sequential in the number of timesteps, because state at step $n+1$ requires state at step $n$. A fixed-point solve is sequential in the number of iterations, and each iteration evaluates all module operators independently. In the experiments below the coupled solve converges in $4$--$10$ Newton iterations on trajectories represented over $1500$ timesteps.

The consequence that cuts deeper is differentiation. Differentiable simulators already exist, and gradients through coupled ODEs are routine: one either unrolls the integrator, at memory linear in the number of steps, or solves the continuous adjoint, at constant memory but still with a backward sweep across the full horizon. Both inherit the sequential structure of the forward march. At a fixed point the gradient is instead the solution of a linear system, $(I - J^\top)\mathbf{v} = \nabla \mathcal{J}$, an object with no time recursion in it at all. It can be solved by Krylov methods whose matrix-vector products are single applications of the module operators. Making those products cheap is precisely what the learned operator buys, and we show below that with an exact integrator in the same role the second-order solver is roughly $100\times$ slower than simple relaxation, which is why the equilibrium view has not been practical before.

The method does not remove time from the representation. Trajectories are still represented on discrete grids; for the oscillator system we represent them in a Chebyshev coefficient basis, which decouples the model's input dimension from the number of timesteps but does not eliminate the time axis. What changes is the outer computational structure: the solver no longer constructs the trajectory by applying a local update rule for every timestep.

This paper makes four contributions. First, it formulates coupled dynamical simulation as a trajectory-level self-consistency problem and situates it explicitly as waveform relaxation with learned subsystem operators. Second, it shows that the classical relaxation iteration is not the right solver for learned operators, and that a Jacobian-free Newton--Krylov method removes the contraction requirement that limits it. Third, it establishes that a single measured quantity---the spectral radius of the learned operator's Jacobian---predicts the convergence boundary of the coupled solve in advance, across both coupling strength and stiffness. Fourth, it demonstrates implicit differentiation through the fixed point, validated three independent ways, including a regime where no other gradient is available.

We do not claim that TWT replaces classical numerical integration. Classical solvers remain the standard tool for accuracy, reliability, and broad applicability, and the systems studied here are small. The purpose of this work is to make a different computational paradigm precise: learn module-level trajectory operators, compose them through self-consistency, and differentiate through the resulting fixed point.

\section{Background and Related Work}

\subsection{Waveform Relaxation}

Waveform relaxation \cite{lelarasmee1982waveform} decomposes a coupled system into subsystems and iterates on whole waveforms: given a current guess for every subsystem's trajectory over $[0,T]$, each subsystem is integrated independently under the inputs those trajectories induce, and the result becomes the next guess. The iteration converges superlinearly on finite windows for Lipschitz systems \cite{miekkala1987convergence}, with a contraction factor that degrades as the window lengthens---which is why practical implementations window the time axis. Parallel-in-time methods such as Parareal pursue the related goal of decoupling sequential depth from timestep count by decomposing the time axis itself \cite{lions2001parareal, gander2015fifty}; TWT instead decomposes by subsystem.

TWT is this algorithm with the inner integration replaced by a learned operator. The consequences go beyond cost. Because the learned operator is differentiable and its Jacobian-vector products cost one network evaluation, solvers that are impractical for classical waveform relaxation become available: we use Newton--Krylov for the forward solve and GMRES for the adjoint, both of which need many Jacobian applications per step.

\subsection{Trajectories as Global Objects}

The view of trajectories as global objects has deep roots in physics. Hamilton's principle selects physical trajectories by extremizing an action functional over a space of possible paths. Variational integrators discretize this principle directly and preserve geometric structure such as symplecticity and momentum maps \cite{marsden2001discrete}. Physics-informed neural networks (PINNs) also adopt a global view by parameterizing a solution function and minimizing differential-equation residuals over a domain \cite{raissi2019physics}.

These approaches differ from TWT in important ways. Variational integrators still construct the trajectory through a discretized temporal scheme. PINNs optimize a new solution for each problem instance and often face ill-conditioned losses, spectral bias, and difficulty balancing residual and boundary terms \cite{krishnapriyan2021characterizing, wang2022and}. TWT instead amortizes subsystem solves into learned trajectory operators and uses self-consistency to compose them at inference time.

\subsection{Neural Surrogates and Neural Operators}

A growing body of work trains neural networks to approximate solution maps for differential equations. Neural ordinary differential equations learn a vector field and integrate it with a numerical solver, often using adjoint methods for memory-efficient differentiation \cite{chen2018neural}. Neural operators such as DeepONet and the Fourier Neural Operator learn mappings between function spaces, allowing full solution fields to be predicted from inputs such as forcing functions or initial conditions \cite{lu2021learning, li2021fourier, kovachki2023neural}.

TWT is closest in spirit to neural operator learning, because each module surrogate maps an input trajectory to an output trajectory. The difference is compositional: TWT uses trajectory operators as modules inside a coupled graph and enforces agreement across modules through a fixed-point solve. The problem is therefore not just learning an isolated solution operator, but composing several learned operators into a mutually consistent coupled simulation. As we show in Section~\ref{sec:contraction}, this makes a property of the operator that isolated-accuracy metrics do not measure---the norm of its Jacobian---directly relevant, and the two properties do not improve together.

\subsection{Differentiable Simulation and Equilibrium Models}

Differentiable simulators allow gradients to propagate through physical dynamics, enabling optimization of controls, parameters, and designs. Frameworks such as DiffTaichi and Jaxley differentiate through simulation programs or biophysical neuron models \cite{hu2020difftaichi, deistler2025jaxley}. These approaches differentiate through the time-stepping computation itself, either by unrolling, which stores intermediate states, or by a continuous adjoint, which integrates backward across the horizon. In both cases the gradient computation inherits the sequential structure of the forward solve.

Deep Equilibrium Models (DEQs) define network outputs as fixed points of a learned transformation and use the implicit function theorem for memory-efficient differentiation \cite{bai2019deep, blondel2022efficient}. TWT borrows the fixed-point view from equilibrium models, but applies it to physical simulation: the fixed point is not a hidden representation of a network layer but a set of subsystem trajectories that mutually satisfy coupling constraints, and---unlike a DEQ---it can be checked against an external ground truth.

\section{Method}

\subsection{Problem Formulation}

Consider a coupled system with $K$ subsystems. The state of subsystem $i$ evolves according to
\begin{equation}
    \dot{\mathbf{x}}_i = f_i(\mathbf{x}_i, \mathbf{u}_i(t), \theta_i), \qquad \mathbf{x}_i(0) = \mathbf{x}_{i,0},
\end{equation}
where $\theta_i$ denotes subsystem parameters and $\mathbf{u}_i(t)$ is a driving input generated by neighboring subsystem states:
\begin{equation}
    \mathbf{u}_i(t) = g_i\left(\{\mathbf{x}_j(t)\}_{j \in \mathcal{N}(i)}\right).
\end{equation}
The coupling structure is represented by a graph $\mathcal{G}=(\mathcal{V},\mathcal{E})$, where nodes are subsystems and edges indicate dependencies.

If the complete driving input $\mathbf{u}_i(t)$ were known for each subsystem, each subsystem could be solved independently. The difficulty is circular: the driving inputs depend on the unknown trajectories.

\subsection{Module-Level Trajectory Surrogates}

For each module type, we train a surrogate $S_\phi$ that approximates the isolated subsystem solution operator:
\begin{equation}
    \hat{\mathbf{x}}_i = S_\phi(\theta_i, \mathbf{u}_i, \mathbf{x}_{i,0}),
\end{equation}
where $\mathbf{u}_i$ is a discretized input trajectory and $\hat{\mathbf{x}}_i$ the predicted output trajectory. The surrogate produces the entire trajectory in one forward evaluation, so time still appears in the representation but inference does not proceed by repeatedly applying a local update rule.

\paragraph{Trajectory representation.} For the oscillator system we represent trajectories by their first $N$ Chebyshev coefficients rather than by raw samples \cite{trefethen2013approximation}. The transform is a fixed linear map in both directions, so it is differentiable and commutes with any linear coupling: $\mathcal{C}(W\mathbf{X}) = W\mathcal{C}(\mathbf{X})$, and coupling can be applied directly in coefficient space. On a $30$-unit window, $256$ coefficients reconstruct van der Pol trajectories to below $0.5\%$ across the full stiffness range studied, against $1500$ raw samples. For the neuron system we use raw samples. We emphasize that the representation is an implementation choice orthogonal to the formulation; the two systems use different ones and the method is unchanged.

\paragraph{Training data.} Data is generated by simulating isolated modules with known driving inputs using a conventional numerical solver. To expose the surrogate to inputs similar to those encountered during coupled simulation, we use a rolling-buffer strategy: early samples are generated from simple driving inputs and stored in a buffer, and later samples draw their inputs from weighted combinations of previously generated output trajectories.

This strategy requires care in two respects that we found to matter more than any architectural choice. First, the buffer can degenerate. Trajectories carrying no signal (a neuron that never fires) produce weaker composed drives, which produce further silent trajectories, and the buffer collapses. We discard degenerate samples before they enter the buffer and verify the written dataset. Second, and more consequentially, the buffer only covers the input distribution it happens to generate. Twice during this work an experiment probed a parameter range the generator had never sampled, and in both cases the resulting error was misattributed to the method before being traced to coverage. We return to this in Section~\ref{sec:limitations}.

The surrogate is a residual MLP with hidden dimension $768$ and six residual blocks ($7.5$M parameters for the oscillator, $8.6$M for the neuron). Input trajectories are concatenated with initial conditions and subsystem parameters and passed through the network, which regresses the full output trajectory. We train with plain MSE; Section~\ref{sec:contraction} reports a negative result on Jacobian regularization.

\subsection{Self-Consistency Solve}

Given trained surrogates, the coupled solution is the fixed point
\begin{equation}
    \mathbf{X} = F(\mathbf{X}), \qquad
    F(\mathbf{X})_i = S_\phi\!\left(\theta_i,\, g_i\!\left(\{\mathbf{X}_j\}_{j \in \mathcal{N}(i)}\right),\, \mathbf{x}_{i,0}\right).
    \label{eq:fixed_point}
\end{equation}
At a zero-residual fixed point, every trajectory equals the one predicted by its module surrogate under the inputs induced by its neighbors.

\paragraph{Solvers and the contraction condition.} The natural iteration is Picard, $\mathbf{X} \leftarrow F(\mathbf{X})$, which is what classical waveform relaxation performs. It converges only while
\begin{equation}
    \rho(J_F) < 1, \qquad J_F = \tfrac{\partial S_\phi}{\partial \mathbf{u}} \cdot W,
    \label{eq:contraction}
\end{equation}
where $W$ is the linear coupling operator; for the all-to-all diffusive coupling used below, $\|W\| = k_c$ exactly. This condition is a property of the iteration, not of the formulation. We therefore also solve Eq.~\ref{eq:fixed_point} by Newton's method applied to $G(\mathbf{X}) = \mathbf{X} - F(\mathbf{X})$, with the linear system handled by GMRES \cite{saad1986gmres}---a Jacobian-free Newton--Krylov method \cite{knoll2004jacobian}. The Jacobian is never formed: each GMRES iteration requires only $(I - J_F)\mathbf{v}$, obtained by one forward-mode differentiation of $F$. Newton converges wherever $I - J_F$ is nonsingular, with no contraction requirement.

\paragraph{Implicit differentiation.} At the converged fixed point $\mathbf{X}^*$, gradients of an objective $\mathcal{J}$ with respect to system parameters $\theta$ follow from the implicit function theorem \cite{bai2019deep, blondel2022efficient}:
\begin{equation}
    \left(I - J_F^\top\right)\mathbf{v} = \nabla_{\mathbf{X}^*}\mathcal{J},
    \qquad
    \frac{d\mathcal{J}}{d\theta} = \mathbf{v}^\top \frac{\partial F}{\partial \theta}.
    \label{eq:ift}
\end{equation}
This linear system carries no dependence on how the forward solve was performed and no time recursion. Expanding $(I - J_F^\top)^{-1}$ as a Neumann series---the standard DEQ backward---reintroduces the condition $\rho(J_F) < 1$, and therefore fails exactly where Newton was introduced to succeed. We solve Eq.~\ref{eq:ift} by GMRES instead, which does not.

\subsection{Exact-Surrogate Interpretation}

Under an idealized exact-surrogate assumption, a zero-residual fixed point of Eq.~\ref{eq:fixed_point} coincides with the unique solution of the original coupled initial value problem, provided the underlying ODE satisfies standard existence and uniqueness conditions. At zero residual every subsystem trajectory equals the solution produced by its governing dynamics under the inputs induced by neighboring trajectories, so the assembled trajectories satisfy the coupled system; by uniqueness they are the physical solution.

For learned surrogates this becomes approximate, and the gap is measurable. We therefore report the self-consistency residual and the trajectory error separately throughout, and additionally report the error obtained by running the same fixed point with an \emph{exact} integrator as the module operator. That third quantity is the floor attributable to the formulation and its discretization; the distance between it and the surrogate result is what learning costs.

The decomposition earned its keep during development. In an early version of this work the module simulator applied the drive at the end of each step while the coupled reference applied it at the start. Both are valid $O(\Delta t)$ discretizations, but they define different fixed points, and the discrepancy, which grows from $0.8\%$ to $10.4\%$ with coupling strength, was indistinguishable from surrogate error until the exact-operator check isolated it. We report both conventions in Section~\ref{sec:validation}.

\section{Experiments}

We study two systems. Coupled van der Pol oscillators, $\ddot{x}_i = \mu(1-x_i^2)\dot{x}_i - k_i x_i + F_i(t)$, are nonlinear with a stable limit cycle and, at large $\mu$, relaxation oscillations; a single surrogate spans $\mu \in [0.5, 5]$, from quasi-harmonic to stiff. Hodgkin--Huxley neurons \cite{hodgkin1952quantitative} coupled through excitatory chemical synapses provide a second system with two properties the oscillators lack: the coupling is nonlinear, since the synaptic current depends on both pre- and postsynaptic state, and the surrogate is a compact model---voltage and gating variables are used to generate data but stay latent, and only the coupling-relevant variable $s(t)$ is predicted.

Surrogate accuracy on held-out data is $3.05\%$ median relative error for the oscillator ($5.24\%$ mean, $16.9\%$ at the $95$th percentile) and $2.61\%$ median for the neuron ($3.35\%$ mean, $5.91\%$ at the $95$th percentile).

\subsection{Validating the Fixed Point}
\label{sec:validation}

Before introducing a surrogate we verify that the fixed point is the right object, by solving Eq.~\ref{eq:fixed_point} with an exact RK4 module operator and comparing against a conventional coupled integration of the same system.

\begin{table}[h]
\centering
\caption{Classical waveform relaxation with an exact module operator, $N=20$ oscillators, $\mu=1$. The fixed point reproduces the coupled reference exactly. The last column shows the same solve under a module discretization that differs from the reference by one step in where the coupling force is sampled: both are valid schemes, but they define different fixed points, and the discrepancy grows with coupling.}
\label{tab:validation}
\begin{tabular}{lccccc}
\toprule
$k_c$ & Feedforward err. & WR err. & WR err. (mismatched) & Picard iters. & Residual \\
\midrule
$0.25$ & $95.6\%$ & $0.0\%$ & $0.8\%$  & $9$  & $1.1\times10^{-12}$ \\
$0.50$ & $147.8\%$ & $0.0\%$ & $1.4\%$  & $13$ & $1.3\times10^{-12}$ \\
$1.00$ & $137.9\%$ & $0.0\%$ & $2.7\%$  & $21$ & $3.1\times10^{-12}$ \\
$1.50$ & $146.5\%$ & $0.0\%$ & $4.0\%$  & $30$ & $2.0\times10^{-12}$ \\
$2.00$ & $140.6\%$ & $0.0\%$ & $10.4\%$ & $40$ & $3.6\times10^{-12}$ \\
\bottomrule
\end{tabular}
\end{table}

The formulation is exact, and the growth of Picard iteration counts with $k_c$ is the contraction condition of Eq.~\ref{eq:contraction} becoming binding.

This baseline also quantifies why the learned operator matters. Solving the same fixed point with Newton rather than Picard, using the exact integrator, takes $1548$\,s at $k_c=2$ against Picard's $13.7$\,s, because every GMRES iteration is a full numerical integration of every subsystem. With the surrogate, the same Jacobian-vector product is one forward-mode pass.

\subsection{Coupled Trajectories}

We solve networks of $N=20$ oscillators with all-to-all diffusive coupling and detuned natural frequencies, sweeping coupling strength $k_c$ and stiffness $\mu$, with five random systems per cell. The feedforward baseline is the same surrogate evaluated with the coupling disabled---the uncoupled view of the same network.

\begin{table}[h]
\centering
\caption{Relative trajectory error against RK4 ground truth, $N=20$, five seeds per cell. Feedforward error is $97$--$153\%$ throughout and is omitted per-cell. Parenthesized entries report seeds converged out of five where not all converged.}
\label{tab:vdp_sweep}
\begin{tabular}{lccccc}
\toprule
& \multicolumn{5}{c}{Coupling strength $k_c$} \\
\cmidrule(lr){2-6}
$\mu$ & $0.25$ & $0.50$ & $1.00$ & $1.50$ & $2.00$ \\
\midrule
$0.5$ & $1.7\%$ & $2.4\%$ & $5.8\%$ & $18.2\%$ & $34.8\%$ (4/5) \\
$1.0$ & $1.0\%$ & $1.5\%$ & $4.0\%$ & $13.8\%$ & $31.9\%$ (3/5) \\
$2.0$ & $1.1\%$ & $1.5\%$ & $3.2\%$ & $8.3\%$  & $31.2\%$ \\
$3.0$ & $1.8\%$ & $2.3\%$ & $4.1\%$ & $8.1\%$  & $16.8\%$ \\
$5.0$ & $3.8\%$ & $4.7\%$ & $7.0\%$ & $10.1\%$ & $15.0\%$ \\
\midrule
Newton iters. & $4$ & $4$--$5$ & $6$--$8$ & $8$--$10$ & $10$--$57$ \\
\bottomrule
\end{tabular}
\end{table}

For $k_c \le 1.5$ every configuration converges on all five seeds, with residuals below $10^{-11}$ and errors between $1.0\%$ and $18.2\%$, against a feedforward baseline that is wrong by more than $100\%$ everywhere. The solve takes $4$--$10$ Newton iterations on trajectories spanning $1500$ timesteps.

\subsection{Contraction Predicts Convergence}
\label{sec:contraction}

\begin{figure}[t]
\centering
\includegraphics[width=0.85\linewidth]{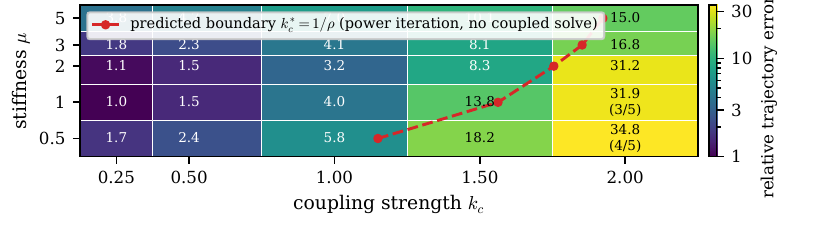}
\caption{The spectral radius of the learned operator, measured by power iteration on the isolated operator, predicts where the coupled solve converges. Cells: relative trajectory error against RK4 ground truth ($N=20$, five seeds per cell; parentheses give seeds converged where not all five did). Line: the predicted boundary $k_c^{*}=1/\rho$. No coupled solve is involved in the prediction.}
\label{fig:contraction}
\end{figure}

The pattern in the last column of Table~\ref{tab:vdp_sweep} is not accidental. We estimate $\rho\big(\partial S_\phi / \partial \mathbf{u}\big)$ by power iteration on the learned operator, using only perturbation probes---no coupled system is involved. Combined with $\|W\| = k_c$, Eq.~\ref{eq:contraction} predicts a convergence boundary $k_c^* = 1/\rho$. Figure~\ref{fig:contraction} overlays the predicted boundary on the observed sweep.

\begin{table}[h]
\centering
\caption{The spectral radius of the learned operator, measured independently of any coupled solve, against the observed behavior at $k_c=2$.}
\label{tab:rho}
\begin{tabular}{lcccc}
\toprule
$\mu$ & $\rho$ (measured) & Predicted $k_c^*$ & Converged at $k_c=2$ & Iters. \\
\midrule
$0.5$ & $0.87$ & $1.15$ & 4/5 & $57$ \\
$1.0$ & $0.64$ & $1.56$ & 3/5 & $45$ \\
$2.0$ & $0.57$ & $1.75$ & 5/5 & $11$ \\
$3.0$ & $0.54$ & $1.85$ & 5/5 & $10$ \\
$5.0$ & $0.52$ & $1.92$ & 5/5 & $10$ \\
\bottomrule
\end{tabular}
\end{table}

The ordering is exact: iteration count, convergence rate, and trajectory error all follow $\rho$ across the stiffness range. The same measurement applied to the true operator gives $\rho \approx 0.54$ at $\mu=1$, so the learned operator is more expansive than the dynamics it approximates, and the coupled solve reaches a lower coupling strength than the physics would allow.

Accuracy and composability, in other words, are different properties, and they do not move together. The surrogate is \emph{least} accurate in isolation at $\mu=5$ ($3.4\%$ against $0.8\%$ at $\mu=1$), yet the coupled solve is \emph{best} there, because contraction dominates once coupling is strong. Across three separately trained surrogates we observe the same tension: better fit accompanied by larger $\rho$.

The obvious remedy does not work naively, either. We trained a surrogate with an added penalty on the operator gain, in the spirit of Jacobian regularization for equilibrium models \cite{bai2021stabilizing}, estimated by random directional probes. The penalty reduced the probed quantity by half and made the coupled solve \emph{worse}. The reason is that a random probe in $n$ dimensions places only $1/n$ of its energy along any one direction, so the estimate is the Frobenius norm rather than the spectral radius; constraining it crushed all directions while leaving the one that governs convergence slightly larger. We report this as a negative result: the correct quantity requires power iteration, not random probing.

\subsection{Implicit Differentiation}

We differentiate a scalar objective with respect to the coupling weights of a ring of oscillators and compare the implicit gradient of Eq.~\ref{eq:ift} against unrolled backpropagation and against central finite differences.

\begin{table}[h]
\centering
\caption{Gradient agreement and memory. Unrolled memory grows with depth $K$; the implicit gradient does not. At coupling $0.5$ the forward Newton solve converges in $14$ iterations, but $\rho > 1$, and only the implicit gradient remains valid.}
\label{tab:ift}
\begin{tabular}{lccc}
\toprule
& Coupling $0.1$, $N=6$ & Coupling $0.1$, $N=50$ & Coupling $0.5$, $N=6$ \\
\midrule
Cosine sim.\ vs.\ unrolled & $1.0000$ & $1.0000$ & $-0.13$ (unrolled diverges) \\
Rel.\ err.\ vs.\ unrolled  & $1.4\times10^{-5}$ & $7.9\times10^{-5}$ & $83.2$ (unrolled diverges) \\
Rel.\ err.\ vs.\ finite diff. & $0.06\%$ & --- & $\mathbf{0.04\%}$ \\
Neumann adjoint & converges & converges & diverges (\texttt{inf}) \\
Unrolled memory, $K{=}2 \to 160$ & $102 \to 146$\,MB & $108 \to 472$\,MB & --- \\
Implicit memory & $112$\,MB & $169$\,MB & $112$\,MB \\
\bottomrule
\end{tabular}
\end{table}

Where the comparison is meaningful the gradients agree to five digits. Where it is not---past the contraction boundary---unrolled backpropagation diverges by a factor of $83$ and the Neumann adjoint returns non-finite values, while the implicit gradient remains correct to $0.04\%$ against finite differences. This is the regime the Newton solver was introduced to reach, and the implicit gradient is the only one that survives it.

To confirm the gradients do useful work in an optimization, we recover the coupling weights of a six-oscillator ring from observed trajectories generated by RK4---the true physics, not the surrogate's own output---by minimizing trajectory mismatch through the fixed point. Over ten random systems, parameter error falls from $83.2\%$ to a median of $1.2\%$ (worst case $2.1\%$) and output error from $105\%$ to $1.3\%$. The implicit and unrolled gradients give identical results at this coupling strength; the implicit one uses memory independent of solver depth.

\subsection{Hodgkin--Huxley Neurons}

An external neuron receives a transient current pulse and drives an internal neuron carrying a self-recurrent synapse, so the feedback loop closes at the internal neuron's own output. We sweep the recurrent weight and measure on the post-pulse window, after the external drive is gone and the recurrent term is all that remains.

\begin{table}[h]
\centering
\caption{Post-pulse relative error and mean activity. WR is the same fixed point with an exact module operator, and is the floor attributable to the formulation; the gap between SC and WR is surrogate error. Feedforward activity is independent of $w_{\mathrm{rec}}$ by construction.}
\label{tab:hh}
\begin{tabular}{lccccccc}
\toprule
& \multicolumn{3}{c}{Error} & \multicolumn{3}{c}{Mean activity} & \\
\cmidrule(lr){2-4} \cmidrule(lr){5-7}
$w_{\mathrm{rec}}$ & FF & SC & WR & GT & FF & SC & Iters. \\
\midrule
$0.5$ & $5.4\%$  & $2.2\%$  & $0.9\%$ & $0.0527$ & $0.0551$ & $0.0527$ & $3$ \\
$1.0$ & $8.2\%$  & $5.8\%$  & $1.0\%$ & $0.0508$ & $0.0551$ & $0.0502$ & $3$ \\
$2.0$ & $17.9\%$ & $9.8\%$  & $1.5\%$ & $0.0474$ & $0.0551$ & $0.0472$ & $4$ \\
$3.0$ & $35.4\%$ & $18.5\%$ & $1.7\%$ & $0.0436$ & $0.0551$ & $0.0439$ & $4$ \\
$4.0$ & $56.3\%$ & $29.0\%$ & $1.9\%$ & $0.0397$ & $0.0551$ & $0.0404$ & $9$ \\
$5.0$ & $54.3\%$ & $56.2\%$ & $1.6\%$ & $0.0405$ & $0.0551$ & $0.0413$ & (no conv.) \\
\bottomrule
\end{tabular}
\end{table}

The activity columns carry the clearer statement. Feedforward predicts that recurrence has no effect at all: its activity is pinned at $0.0551$ regardless of $w_{\mathrm{rec}}$. The true activity falls monotonically from $0.0527$ to $0.0397$ as recurrence strengthens---a consequence of the synaptic reversal potential, which makes strong input depolarizing at rest but shunting during a spike---and the fixed point tracks that dependence where the uncoupled view, by construction, cannot.

The error columns are more equivocal, and we report them plainly. Self-consistency roughly halves feedforward error, but the exact-operator floor sits near $1.9\%$, so most of the remaining error is the surrogate rather than the formulation. At $w_{\mathrm{rec}}=5$ the solve does not converge for this surrogate. The neuron surrogate is comparably accurate to the oscillator one in isolation, so the gap reflects amplification through a fixed point whose conditioning is worse, consistent with Section~\ref{sec:contraction}.

\section{Discussion}

The experiments support the premise: coupled simulation can be reformulated as trajectory-level self-consistency among learned module operators, and the reformulation changes the structure of both the forward computation and its derivative. The validation in Section~\ref{sec:validation} establishes that the fixed point is the correct object independently of any learning, and the exact-operator floor reported throughout separates what the formulation achieves from what the surrogate costs.

Two findings we did not anticipate seem worth emphasizing.

The first is that the classical relaxation iteration is the wrong solver for learned operators, and that this is not a detail. Picard is bound by $\rho(\partial S_\phi/\partial \mathbf{u})\,\|W\| < 1$, and a learned operator is typically more expansive than the dynamics it approximates, so the bound binds earlier than the physics requires. Newton removes the condition entirely, and it is affordable only because the operator is learned: with an exact integrator in the same role it is roughly $100\times$ slower than Picard. The equilibrium view and the learned operator are not independent choices; each is what makes the other practical.

The second is that operator accuracy and operator composability are distinct properties that can be measured separately and that do not improve together. A surrogate can fit better in isolation and compose worse, and we observe this across three independently trained models and across stiffness within a single model. Standard surrogate evaluation reports only the first property. For any method that composes learned operators, the second appears to be equally predictive of end-to-end behavior, and---via power iteration on the Jacobian---it can be measured before any coupled solve is attempted.

The results also clarify what the method does \emph{not} solve. TWT does not remove the need to represent trajectories, nor does it eliminate approximation error. The self-consistency residual measures agreement under the learned surrogate, not satisfaction of the original differential equation, so a residual of $10^{-12}$ can coexist with substantial physical error---as it does at $k_c=2$. Reporting both, together with the exact-operator floor, is necessary for the numbers to be interpretable.

\section{Limitations and Future Work}
\label{sec:limitations}

The present results are an early demonstration rather than a mature alternative to classical solvers. The systems studied here are small: $20$ coupled oscillators and neuron pairs. Nothing in the formulation is specific to that scale, but we have not demonstrated it beyond it, and the structural claims about sequential depth would be considerably more convincing at $10^3$ modules. Surrogate error of $1$--$4\%$ is likewise far from solver-grade, and fixed-point conditioning amplifies it: a converged fixed point is only as meaningful as the operators from which it is built. The exact-operator floors reported here ($0.0\%$ for oscillators, $1.9\%$ for neurons) bound what better surrogates could achieve within the same formulation.

For the same reason we report sequential depth---$4$--$10$ iterations against $1500$ timesteps---and not wall-clock speedup. Neither our implementation nor the reference integrator is optimized, so the comparison would not be meaningful. The structural claim is that iteration count is not tied to trajectory length and that iterations are parallel across modules; the practical claim requires an optimized implementation we have not built. For reference, all reported experiments were trained and run in approximately one day on a single rented NVIDIA RTX~5090 (32\,GB VRAM) instance with a 16-core AMD Ryzen~7 7800X3D host, roughly $24$ GPU-hours in total; preliminary experimentation over the project's duration used additional heterogeneous rented GPUs.

The dominant practical failure mode was training-distribution coverage. Twice, an experiment exercised a parameter range the data generator had never sampled (once in drive magnitude, once in recurrent weight), and in both cases the error was initially misattributed to the method. Composing learned operators makes this worse than in standard surrogate use, because the fixed-point iteration itself determines the inputs the operator sees, and those inputs are not known in advance. Generating training data that covers the distribution induced by coupling, rather than a distribution chosen beforehand, seems to us a central unsolved problem for this class of method.

Finally, the experiments cover ODE systems with linear or synaptic coupling; PDEs, stiff multiscale systems, chaotic dynamics, discontinuities, and event-driven systems require separate investigation. Windowing the time axis, the classical remedy for slow waveform relaxation convergence, would also reduce $\rho$ directly and is a natural next step we did not pursue. Beyond that, the most promising directions we see are learned preconditioners for the fixed-point solve, compressed or latent trajectory representations for long horizons, and training objectives that control the spectral radius directly rather than through the proxy we found insufficient.

\section{Conclusion}

Time Without Timesteps reframes coupled dynamical simulation as a fixed-point problem over trajectories, following classical waveform relaxation but replacing the inner numerical integration with learned trajectory operators. The reformulation changes the sequential structure of the forward solve, converging in $4$--$10$ iterations on trajectories spanning $1500$ timesteps, and it changes the structure of the gradient, replacing a backward march through the integrator with a linear system that can be solved at memory independent of solver depth. We show that the convergence boundary of the coupled solve is predicted in advance by a single measured property of the learned operator, that this property is distinct from isolated accuracy and does not improve alongside it, and that past the boundary the implicit gradient is the only one that remains valid. The result is not a universal alternative to numerical integration, but a computational paradigm for differentiable, modular simulation whose failure modes we can now measure rather than merely observe.

\bibliographystyle{plain}
\bibliography{references}

\begin{thebibliography}{10}

\bibitem{bai2019deep}
Shaojie Bai, J.~Zico Kolter, and Vladlen Koltun.
\newblock Deep equilibrium models.
\newblock In {\em Advances in Neural Information Processing Systems},
  volume~32, 2019.

\bibitem{bai2021stabilizing}
Shaojie Bai, Vladlen Koltun, and J.~Zico Kolter.
\newblock Stabilizing equilibrium models by {J}acobian regularization.
\newblock In {\em International Conference on Machine Learning}, 2021.

\bibitem{blondel2022efficient}
Mathieu Blondel, Quentin Berthet, Marco Cuturi, Roy Frostig, Stephan Hoyer,
  Felipe Llinares-L{\'o}pez, Fabian Pedregosa, and Jean-Philippe Vert.
\newblock Efficient and modular implicit differentiation.
\newblock In {\em Advances in Neural Information Processing Systems},
  volume~35, 2022.

\bibitem{chen2018neural}
Ricky T.~Q. Chen, Yulia Rubanova, Jesse Bettencourt, and David Duvenaud.
\newblock Neural ordinary differential equations.
\newblock In {\em Advances in Neural Information Processing Systems},
  volume~31, 2018.

\bibitem{deistler2025jaxley}
Michael Deistler, Kyra~L. Kadhim, Matthijs Pals, Jonas Beck, Ziwei Huang,
  Manuel Gloeckler, Janne~K. Lappalainen, Cornelius Schr{\"o}der, Philipp
  Berens, Pedro~J. Gon{\c{c}}alves, and Jakob~H. Macke.
\newblock Jaxley: Differentiable simulation enables large-scale training of
  detailed biophysical models of neural dynamics.
\newblock {\em Nature Methods}, 22(12):2649--2657, 2025.

\bibitem{gander2015fifty}
Martin~J. Gander.
\newblock 50 years of time parallel time integration.
\newblock In {\em Multiple Shooting and Time Domain Decomposition Methods},
  pages 69--113. Springer, 2015.

\bibitem{hodgkin1952quantitative}
Alan~L. Hodgkin and Andrew~F. Huxley.
\newblock A quantitative description of membrane current and its application to
  conduction and excitation in nerve.
\newblock {\em The Journal of Physiology}, 117(4):500--544, 1952.

\bibitem{hu2020difftaichi}
Yuanming Hu, Luke Anderson, Tzu-Mao Li, Qi~Sun, Nathan Carr, Jonathan
  Ragan-Kelley, and Fr{\'e}do Durand.
\newblock {DiffTaichi}: Differentiable programming for physical simulation.
\newblock In {\em International Conference on Learning Representations}, 2020.

\bibitem{knoll2004jacobian}
Dana~A. Knoll and David~E. Keyes.
\newblock Jacobian-free {N}ewton--{K}rylov methods: A survey of approaches and
  applications.
\newblock {\em Journal of Computational Physics}, 193(2):357--397, 2004.

\bibitem{kovachki2023neural}
Nikola Kovachki, Zongyi Li, Burigede Liu, Kamyar Azizzadenesheli, Kaushik
  Bhattacharya, Andrew Stuart, and Anima Anandkumar.
\newblock Neural operator: Learning maps between function spaces with
  applications to {PDE}s.
\newblock {\em Journal of Machine Learning Research}, 24(89):1--97, 2023.

\bibitem{krishnapriyan2021characterizing}
Aditi~S. Krishnapriyan, Amir Gholami, Shandian Zhe, Robert~M. Kirby, and
  Michael~W. Mahoney.
\newblock Characterizing possible failure modes in physics-informed neural
  networks.
\newblock In {\em Advances in Neural Information Processing Systems},
  volume~34, 2021.

\bibitem{lelarasmee1982waveform}
Ekachai Lelarasmee, Albert~E. Ruehli, and Alberto~L. Sangiovanni-Vincentelli.
\newblock The waveform relaxation method for time-domain analysis of large
  scale integrated circuits.
\newblock {\em IEEE Transactions on Computer-Aided Design of Integrated
  Circuits and Systems}, 1(3):131--145, 1982.

\bibitem{li2021fourier}
Zongyi Li, Nikola Kovachki, Kamyar Azizzadenesheli, Burigede Liu, Kaushik
  Bhattacharya, Andrew Stuart, and Anima Anandkumar.
\newblock Fourier neural operator for parametric partial differential
  equations.
\newblock In {\em International Conference on Learning Representations}, 2021.

\bibitem{lions2001parareal}
Jacques-Louis Lions, Yvon Maday, and Gabriel Turinici.
\newblock R{\'e}solution d'{EDP} par un sch{\'e}ma en temps ``parar{\'e}el''.
\newblock {\em Comptes Rendus de l'Acad{\'e}mie des Sciences -- Series I --
  Mathematics}, 332(7):661--668, 2001.

\bibitem{lu2021learning}
Lu~Lu, Pengzhan Jin, Guofei Pang, Zhongqiang Zhang, and George~Em Karniadakis.
\newblock Learning nonlinear operators via {DeepONet} based on the universal
  approximation theorem of operators.
\newblock {\em Nature Machine Intelligence}, 3(3):218--229, 2021.

\bibitem{marsden2001discrete}
Jerrold~E. Marsden and Matthew West.
\newblock Discrete mechanics and variational integrators.
\newblock {\em Acta Numerica}, 10:357--514, 2001.

\bibitem{miekkala1987convergence}
Ulla Miekkala and Olavi Nevanlinna.
\newblock Convergence of dynamic iteration methods for initial value problems.
\newblock {\em SIAM Journal on Scientific and Statistical Computing},
  8(4):459--482, 1987.

\bibitem{raissi2019physics}
Maziar Raissi, Paris Perdikaris, and George~E. Karniadakis.
\newblock Physics-informed neural networks: A deep learning framework for
  solving forward and inverse problems involving nonlinear partial differential
  equations.
\newblock {\em Journal of Computational Physics}, 378:686--707, 2019.

\bibitem{saad1986gmres}
Youcef Saad and Martin~H. Schultz.
\newblock {GMRES}: A generalized minimal residual algorithm for solving
  nonsymmetric linear systems.
\newblock {\em SIAM Journal on Scientific and Statistical Computing},
  7(3):856--869, 1986.

\bibitem{trefethen2013approximation}
Lloyd~N. Trefethen.
\newblock {\em Approximation Theory and Approximation Practice}.
\newblock SIAM, 2013.

\bibitem{wang2022and}
Sifan Wang, Xinling Yu, and Paris Perdikaris.
\newblock When and why {PINN}s fail to train: A neural tangent kernel
  perspective.
\newblock {\em Journal of Computational Physics}, 449:110768, 2022.

\end{thebibliography}
\end{document}